\documentclass{article} 
\usepackage[final]{colm2026_conference}

\usepackage{microtype}
\usepackage{hyperref}
\usepackage{url}
\usepackage{amsmath}
\usepackage{booktabs}
\usepackage{graphicx}
\usepackage{subcaption}
\usepackage[table]{xcolor} 
\usepackage{tikz}
\usetikzlibrary{arrows.meta, positioning, calc}
\usepackage{placeins} 
\usepackage{float} 

\renewcommand{\phi}{\varphi}
\newcommand{\pmval}[2]{$#1{\scriptstyle\,\pm#2}$}
\definecolor{ourshade}{gray}{0.9} 

\usepackage{lineno}

\definecolor{darkblue}{rgb}{0, 0, 0.5}
\hypersetup{colorlinks=true, citecolor=darkblue, linkcolor=darkblue, urlcolor=darkblue}

\title{Inverse Reasoning Models for Efficient Post-Training and Verification from Answer-Only Data}

\author{Eric O.~Korman \\ Striveworks \\ \texttt{eric.korman@gmail.com}}

\begin{document}

\ifcolmsubmission
\linenumbers
\fi

\maketitle

\begin{abstract}
Post-training reasoning large language models (LLMs) often requires making a tradeoff between costly supervision and compute inefficiency: supervised finetuning typically requires reasoning traces, which is data inefficient since in many applications one only has prompt-answer tuples and lacks the groundtruth chain-of-thought (CoT) sequence. On the other hand, reinforcement learning (with verifiable rewards, for example) can operate without CoT data but it is much more compute-inefficient since it requires multiple autoregressive completions per optimization step, none of which are guaranteed to give a non-zero policy gradient. In this work we explore a middleground, a self-distillation technique that uses answer-only data, treating CoT as latent variables. This relies on training a partner ``inverse reasoning LLM" that generates the original's reasoning traces conditioned on the answer. The inverse model is trained once and reused across incoming batches of answer-only data, with all generation happening offline and up front, making the approach particularly suited to continual post-training. Using Qwen3-0.6B on the Countdown and GSM8K tasks, we show that after GRPO training has saturated, a single offline distillation round from the inverse model roughly doubles pass@1 on hard prompts (e.g.\ $0.08 \to 0.19$ on Countdown) and, on Countdown, outperforms rejection-sampling self-distillation on every hard-tail metric at every mix ratio (with smaller gains on GSM8K); rejection sampling by construction has no signal on problems the model never solves on its own. Continued GRPO on the fresh data retains an edge at hard pass@1 ($0.23$) but trails distillation at pass@8 and pass@32 while regenerating rollouts at every step. As an added bonus we see that the inverse model has a second application as an accurate verifier of correctness. When used for best-of-$n$ selection at test time, it raises selection accuracy on Countdown from 0.753 (majority voting) to 0.969.
\end{abstract}

\section{Introduction}

Chain-of-thought (CoT) reasoning \citep{wei2023chainofthoughtpromptingelicitsreasoning} has extended the abilities of large language models (LLMs) from their initial defining task of text generation to being capable of solving complex problems. From the perspective of an LLM, there is no difference between the tokens that are part of the prompt, the CoT sequence, or the final answer. Functionally, however, these tokens serve very different purposes: CoT tokens are largely a means to an end, providing additional computation steps that bridge the prompt and the final answer. The asymmetry also has a counterpart at the data level, where question–answer pairs are often abundant but reasoning traces are scarce. This has been a large factor in the focus on reinforcement learning (RL) techniques such as GRPO \citep{shao2024deepseekmath} over supervised finetuning (SFT), which would require groundtruth reasoning traces, especially on verifiable tasks \citep{lambert2024tulu}. Here, groundtruth CoT is not needed as the learning signal comes solely from whether the final answer is correct. While these approaches have proven effective, they are typically far less compute-efficient than supervised finetuning, requiring many rollouts per prompt before the LLM stumbles upon a correct answer that provides a learning signal.

The starting point for our work is how we can do efficient, supervised finetuning for CoT reasoning LLMs on question-answer pairs without reasoning data. We do this by treating the CoT sequences as latent variables and take a variational inference approach that makes use of a learned \textit{inverse model}. This is a separate LLM trained to generate the CoT of the original LLM given both the prompt and answer. Using an auxiliary answer-conditioned model has firm standing in the literature, including the earlier work of \citet{zelikman2022star} and  \citet{phan2023training}, and the very recently developed techniques of RAVR \citep{lin-etal-2026-ravr} and CoVRL \citep{wen2025coupled}, though these operate in the online RL setting rather than the SFT/offline-RL setting we target (see Section~\ref{sec:related} for a detailed comparison).

Our main experiments are self-distillation applications, where the inverse model is trained from CoT on correct answers from the forward model, and then the forward model is finetuned so its CoT matches the inverse model's. This works with either the original data the inverse model was trained on or on new data, which can be considered an instance of offline reinforcement learning. 

As an additional use case, we find that the inverse model forms an accurate verifier of correctness, which we show can be used as a more efficient alternative to (adaptive) best-of-n sampling \citep{cobbe2021training, aggarwal2023let}. 

For validating our techniques, we run experiments using Qwen3-0.6B \citep{yang2025qwen3} with the Countdown \citep{gandhi2024stream} and GSM8K \citep{cobbe2021training} datasets. In both environments we start from a model whose GRPO training has saturated and show that distillation from the inverse model roughly doubles pass@1 on \emph{hard} prompts (those the starting model solves at most 25\% of the time) while maintaining overall pass@8 and pass@32. On Countdown it beats the rejection-sampling baseline on every hard-tail metric at every mix ratio, and on GSM8K it matches or modestly exceeds it; the hard tail is exactly the regime where rejection sampling is structurally blind since it can only train on problems the forward model already solves. On the other hand, the inverse model can propose a correct-leading CoT for problems the forward model never solves on its own. Continued GRPO on fresh data remains the strongest method at hard pass@1, but distillation surpasses it at pass@8 and pass@32 without any on-policy rollouts during training.

We open-source the codebase for all of our experiments at \url{https://github.com/ekorman/dialectic}.

\section{Method}

\subsection{Overview}
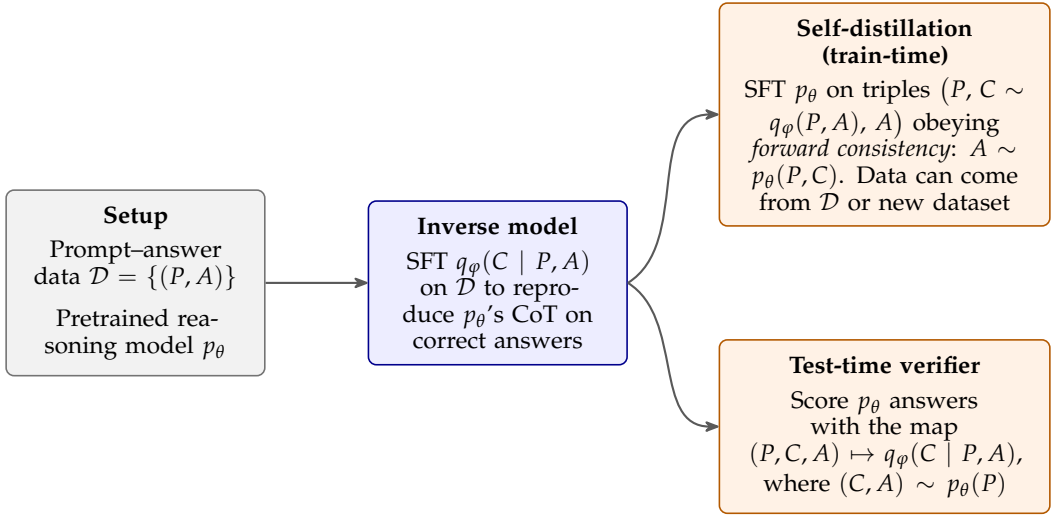
\begin{figure}[t]
\centering
\begin{tikzpicture}[
    >={Stealth[round]},
    font=\small,
    base/.style={draw, rounded corners=3pt, align=center, inner sep=6pt,
                 line width=0.6pt},
    setupbox/.style={base, fill=black!5,   draw=black!55,      text width=3.0cm},
    corebox/.style={base,  fill=blue!7,    draw=blue!55!black, text width=3.0cm},
    appbox/.style={base,   fill=orange!10, draw=orange!70!black, text width=3.95cm},
    arrow/.style={->, line width=0.8pt, draw=black!65},
]

\node[setupbox] (setup)
    {\textbf{Setup}\\[3pt]
     Prompt--answer data $\mathcal{D}=\{(P,A)\}$\\[6pt]
     Pretrained reasoning model $p_\theta$};

\node[corebox, right=1.35cm of setup] (inv)
    {\textbf{Inverse model}\\[3pt]
     SFT $q_\phi(C\mid P,A)$ on $\mathcal D$ to reproduce $p_\theta$'s CoT on correct answers};

\node[appbox, above right=-0.35cm and 1.2cm of inv] (sft)
    {\textbf{Self-distillation (train-time)}\\[3pt]
     SFT $p_\theta$ on triples $\big(P,\,C\sim q_\phi(P,A),\,A\big)$
     obeying \textit{forward consistency}: $A\sim p_\theta(P,C)$.
     Data can come from $\mathcal D$ or new dataset};

\node[appbox, below right=-0.35cm and 1.2cm of inv] (ver)
    {\textbf{Test-time verifier}\\[3pt]
     Score $p_\theta$ answers with the map\\
     $(P,C,A)\mapsto q_\phi(C\mid P,A)$,\\
     where $(C,A)\sim p_\theta(P)$};

\draw[arrow] (setup) -- (inv);
\draw[arrow] (inv.east) to[out=32, in=180] (sft.west);
\draw[arrow] (inv.east) to[out=-32, in=180] (ver.west);

\end{tikzpicture}
\caption{Overview of our method. Starting from answer-only data $\mathcal{D}$ and a
reasoning model $p_\theta$ that already solves part of it, we train an \emph{inverse
model} $q_\phi$ to reconstruct $p_\theta$'s chain-of-thought from the answer. The inverse
model then supports two downstream uses: (top) self-distillation, where $p_\theta$ is
finetuned on $q_\phi$-generated reasoning traces that remain forward-consistent with
$p_\theta$, after which the finetuned $p_\theta$ is deployed alone; and (bottom) test-time
verification, where $q_\phi$ scores candidate answers sampled from $p_\theta$.}
\label{fig:overview}
\end{figure}
We now give an overview of our method, whose summary can be seen in Figure \ref{fig:overview}. Consider a dataset distribution $\mathcal D =  \{ (P, A)\} \subset \mathcal P \times \mathcal A$ where $\mathcal P$ is a space of prompts and $\mathcal A$ is a space of answers.  Let $p_\theta$ be an LLM with parameters $\theta$ (we will also use $p_\theta$ to denote the probability distribution the LLM determines on the token space). We assume that the LLM is a reasoning model, i.e. from a prompt $P$ an answer is arrived at by first generating a CoT sequence $C$ (commonly distinguished from the prompt and answer by being enclosed in \texttt{<think>} tags) from $P$ and then generating $A$ from $P$ and $C$.

Our ultimate goal is to treat the CoT sequences as latent variables and find $\theta$ that maximizes the log-likelihood 
\[
\ell(\theta) = \sum_{(P, A) \in \mathcal D} \log p_\theta(A | P) 
\]

This is difficult because we only readily have access to $p_\theta(C | P)$ and $p_\theta(A | C, P)$, and computing $p_\theta(A | P)$ by marginalizing over $C$ is intractable. However, we can take advantage of Fisher's identity \citep{cappe2005inference} 
\begin{equation}
\nabla_\theta \ell(\theta_0) = \sum_{(P, A) \in \mathcal D} \mathbb E_{C \sim p_{\theta_0} (\cdot | P, A)} \left.\left[\nabla_\theta \log p_\theta(C, A | P)\right]\right\vert_{\theta = \theta_0} \label{eq:fisher}
\end{equation}
if we can get a hold of the posterior $p_{\theta} (C | P, A)$. For this we take a variational inference approach and approximate this distribution via a separate language model $q_\phi$ that generates $C$ from $(P, A)$, writing $q_\phi(\cdot \mid P, A) \approx p_\theta(\cdot \mid P, A)$.  This is trained via supervised finetuning on the completions that $p_\theta$ got correct.

\subsection{Training $q_\phi$} \label{sec:training-q}
Naively training $q_\phi$ via gradient descent against the negative log likelihood (NLL) objective on correct completions,
\begin{equation}
L_{nll}(P) = -\log q_\phi(C | P, A), ~ \text{ where } C \sim p_\theta(P), A \sim p_\theta(P, C), (P, A) \in \mathcal D \label{eq:nll}
\end{equation}
turns out to not work well in practice: we found that $q_\phi$ trained in this way tends to ignore $A$ and will generate some reasoning traces that lead to a wrong answer. This is likely due to the token imbalance (of most environments, including the ones we use) where CoT sequences are significantly longer than the answer sequences. Therefore, to encourage $q_\phi$ to take $A$ into account, we supplement the NLL loss with a contrastive term. We do this by picking a group size $g > 1$ and generate $g$-many completions from $p_\theta$. We then filter the generated group completions to those where the group contains at least one correct and one incorrect answer. Let $\ell_+(P)$ and $\ell_-(P)$ be the average log probs (across tokens) that $q_\phi$ assigns for all completions of a prompt $P$ that lead to correct and incorrect answers, respectively. We then define the contrastive margin loss as
\begin{equation}
L_{cont}(P) = \max\left(m - (\ell_+(P) - \ell_-(P)) , 0\right) \label{eq:contrastive}
\end{equation}
where $m > 0$ is the \textit{margin} hyperparameter. We combine these, via a hyperparameter $\lambda$, to form our total objective
\begin{equation}
L_q = L_{nll} + \lambda L_{cont}, \label{eq:q-loss}
\end{equation}
which we optimize over minibatches of prompts using stochastic gradient descent. Note that the $L_{nll}$ term only takes in correct completions while $L_{cont}$ makes use of the entire group of completions.

\subsection{Applying $q_\phi$}
From $q_\phi$ we get two applications.

\subsubsection{Distillation from $q_\phi$ to $p_\theta$} \label{sec:distill-q-to-p-method}
The first, which is our main focus, finetunes $p_\theta$ from the CoT traces generated by $q_\phi$. This proceeds by starting from a dataset $\mathcal D'$ (which may or may not be $\mathcal D$) of prompt-answer pairs, and then using $q_\phi$ to generate CoT traces for it. From these, we filter to include only those that lead $p_\theta$ to the correct answer, thus forming the dataset,

\[
\mathcal D'_{q_\phi, p_\theta} = \{ (P, C, A) \mid C \sim q_\phi(P, A), ~ A \sim p_\theta(P, C), ~ (P, A) \in \mathcal D' \}.
\]

which we can directly do SFT of $p_\theta$ on since this has complete reasoning traces with answers. Concretely, we sample a single CoT from $q_\phi$ per prompt and keep it only if it is \emph{forward consistent}: conditioned on the prompt and $q_\phi$'s CoT, $p_\theta$ greedy-decodes the correct answer. We additionally drop $q_\phi$-CoTs for prompts on which every $p_\theta$ rollout is already correct, concentrating $q_\phi$'s supervision where $p_\theta$ needs it.

As the optimization process updates the weights of $p_\theta$, the distributional approximation accuracy of $q_\phi$ may drift from the changing $p_\theta(\cdot | P, A)$ since $\mathcal D'_{q_\phi, p_\theta}$ is fixed from the start (we do not resample from $p_\theta$ during distillation). Therefore as a regularization term to keep $p_\theta$ from drifting too much we supplement $\mathcal D'_{q_\phi, p_\theta}$ with rejection sampling trajectories of $p_\theta$ alone, the set of which we will denote by $\mathcal D'_{RS_{p_\theta}}$: we sample a group of completions from $p_\theta$ for each prompt and keep every correct completion as its own training example. We let the parameter $0 \le \mu \le 1$, which we call the \textit{mix ratio}, denote the ratio of data in a batch that is drawn from $\mathcal D'_{q_\phi, p_\theta}$ (the rest coming from $\mathcal D'_{RS_{p_\theta}}$), with examples drawn uniformly from each pool. We experiment and show the tradeoffs for choosing values of $\mu \in \{0, 0.25, 0.5, 0.75, 1\}$.

We note importantly that in this application $q_\phi$ is used only during training: at deployment the finetuned $p_\theta$ operates alone, leaving inference cost and interface unchanged.

We also highlight where the generation cost is paid. All sampling, namely the CoT traces from $q_\phi$ and the rejection-sampling trajectories from $p_\theta$, happens once, up front, before any weight update. The remainder of training is plain SFT over a fixed dataset. This is in contrast to online policy-gradient methods such as GRPO, which must regenerate a fresh group of rollouts for every prompt at every training step because the rollouts must come from the current policy. Our approach is thus naturally an offline one: the up-front generations can be batched with high throughput, reused across runs (e.g. across the mix-ratio sweep in Section \ref{sec:distill-q-to-p}), and the training itself never blocks on autoregressive sampling.

\subsubsection{Using $q_\phi$ as a verifier of $p_\theta$}
The second application uses $q_\phi$ as a verifier of $p_\theta$'s correctness. Given a prompt $P$, CoT $C$, and answer $A$ computed from $p_\theta$, we can use $q_\phi$ to compute $q_\phi(C | P, A)$, which is its estimate for the likelihood of the reasoning given the prompt and generated answer. Concretely, the score we use is the length-normalized log-likelihood $\frac{1}{|C|+1} \log q_\phi(C, \texttt{eos} \mid P, A)$, computed in a single forward pass; since attention among CoT tokens is causal (Section \ref{sec:experiments}), this is an ordinary autoregressive sequence likelihood. Unlike the distillation application, here $q_\phi$ is deployed at test time; this requires no groundtruth answer since $q_\phi$ scores the answer that $p_\theta$ itself generated. Because of the contrastive term (eq. \ref{eq:contrastive}) used in training $q_\phi$, it is reasonable to expect that this number correlates with the correctness of $A$. In other words, we can think of the function $(P, C, A) \mapsto q_\phi(C | P, A)$  as similar to the trained verifiers of \citet{cobbe2021training}. In Section \ref{sec:q-verifier-exp} we measure its performance as both a best-of-n and adaptive \citep{aggarwal2023let} sampler.

\FloatBarrier

\section{Comparison to related work} \label{sec:related}

RAVR \citep{lin-etal-2026-ravr} and CoVRL \citep{wen2025coupled}, which also take a variational inference approach, are most similar to our work but their focus is on RL instead of SFT: they use the answer-conditioned model to improve the sample efficiency of online algorithms such as GRPO, 
and still require continuously sampling multiple rollouts per prompt from the current policy during training. Because of our contrastive loss (eq. \ref{eq:contrastive}), our method also requires multiple rollouts per prompt for training $q_\phi$, but this is incurred once up-front and the same $q_\phi$ can be used across fresh datasets as we show in Section \ref{sec:distill-q-to-p}. Consistent with their findings, our results in Section \ref{sec:fcr-exp} show that $q_\phi$ allows for more usable traces per generation compared to sampling from $p_\theta$ alone. We also highlight that both RAVR and CoVRL use the same LLM for both the forward and inverse model, using different prompt templates to switch between the two modes. We, on the other hand, train a separate inverse model, which as we show additionally serves as an accurate verifier for the forward model (Section \ref{sec:q-verifier-exp}).

STaR \citep{zelikman2022star} and TRICE \citep{phan2023training} are early works that used rejection sampling and answer-conditioned reasoning. STaR is an iterative finetuning algorithm built on rejection sampling and \textit{rationalization}, which provides reasoning conditioned on the groundtruth answer, while TRICE treats CoT as a latent variable as we do. Neither trains a separate inverse model to get an amortized posterior: STaR does in-context proposals by injecting the answer into the prompt, and TRICE utilizes a non-amortized Markov-chain Monte Carlo expectation-maximization algorithm with a memory bank.

From a data-efficiency perspective, we view our work as a middle ground between GRPO-style reinforcement learning (e.g. RAVR/CoVRL) and the finetuning technique of STaR. The former techniques are able to take advantage of rollouts that do not get the correct answer (indeed, to get a learning signal from GRPO on the binary tasks we have been considering requires groups that have both correct and incorrect answers), while pure SFT does not. In our approach, these mixed rollouts are essential for the contrastive term that is needed to effectively train $q_\phi$.

\section{Experiments and Results} \label{sec:experiments}

We use two environments: GSM8K \citep{cobbe2021training} and Countdown \citep{gandhi2024stream}. We recall that the Countdown task involves finding an arithmetic expression that uses $n$ given numbers and evaluates to a target number. We programmatically generate the Countdown datasets we use, taking $n=4$ and generating 50,000 examples. We then break this into an 80/10/10 train/val/test split. For GSM8K we use the official train/test split, with no validation set carved out, and use the test set only for final reporting.

For each environment, we begin with the pre-trained Qwen3-0.6B model \citep{yang2025qwen3} and train it via GRPO on the train dataset until performance saturates, forming the forward model $p_\theta$ for all experiments. Our interest is in what can be extracted \emph{after} RL has plateaued, when further GRPO yields diminishing returns. We then train $q_\phi$ on this dataset as described in Section \ref{sec:training-q}.

While there are no real restrictions on $q_\phi$ architecturally, in our experiments we take $q_\phi$ to be LoRA \citep{hu2022lora}-finetuned from a copy of $p_\theta$. Separately, we drop the causality restriction on the attention layers within the prefix, which is the concatenation $[P; A]$. Explicitly, all tokens within $[P; A]$ attend to one another, while the CoT generation is causal, i.e. those tokens attend to $[P; A]$ and the preceding CoT tokens. The same mask is used when training $q_\phi$ and when scoring with it as a verifier.

Full details of our training datasets and hyperparameters can be found in Appendix \ref{appendix:training-details}.

\subsection{FCR of $q_\phi$} \label{sec:fcr-exp}
The core metric we use to evaluate $q_\phi$ is the \textit{forward consistency rate at n ($\text{FCR@}n$)}. This is the proportion of prompts where $p_\theta$ generates a correct answer out of $n$ completions when conditioned on the CoT generated by $q_\phi$ (the filter used when building training data, described in Section \ref{sec:distill-q-to-p-method}, is the greedy variant of this check).
Of course, this alone is not a complete measure of $q_\phi$: if $q_\phi$ just repeated the answer it is conditioned on then this makes $p_\theta$'s job easy and FCR will be high. 
However, this metric is important since when using $q_\phi$ to finetune $p_\theta$ we only use those CoT that are forward consistent. In Table \ref{table:fcr} we report the FCR scores for the inverse models trained on each environment, at various values of $n$. To show that $q_\phi$ can extend the ability of $p_\theta$, we also report FCR on the subset of prompts where $p_\theta$ by itself generates the correct answer at most 25\% of the time, which we label as \textit{FCR-hard}.

The results for Countdown and GSM8K are in Table \ref{table:fcr}. We see that $q_\phi$ drastically increases the set of solvable problems in Countdown, while for GSM8K the gains are concentrated on the hard subset (and are significant, though not as extreme). We believe one of the main reasons for the discrepancy is that an answer in Countdown, being a full arithmetic expression, exposes much more of the solution's structure to reason backwards from than the single number that constitutes a GSM8K answer. This does not make $q_\phi$ vacuous on Countdown: its reasoning is trained to match $p_\theta$'s and, crucially, it transfers: as we see in the following distillation experiments, finetuning $p_\theta$ on it gives a significant improvement. Appendix \ref{appendix:qualitative} shows a qualitative example of this on a prompt that $p_\theta$ never solves on its own: $q_\phi$'s CoT is a non-trivial backward decomposition of the answer, and after distillation $p_\theta$ reproduces the same strategy unaided. A second factor is that GSM8K simply has less room for improvement, with the baseline $p_\theta$ already at 0.94 pass@32. For our main application of distilling $q_\phi$ into $p_\theta$, what matters most is the high FCR on hard prompts, i.e. extending the set of prompts from which a correct-leading CoT can be obtained at all. When we run SFT with mix ratio $\mu > 0$ we sample CoT from both $p_\theta$ and $q_\phi$, so prompts that $p_\theta$ solves but that are not forward consistent under $q_\phi$ are not lost.

\begin{table}[t]
  \centering
  \begin{tabular}{lcccccc}
\toprule
& \multicolumn{3}{c}{Countdown} & \multicolumn{3}{c}{GSM8K} \\
\cmidrule(lr){2-4} \cmidrule(lr){5-7}
& FCR@1 & FCR@8 & FCR@32 & FCR@1 & FCR@8 & FCR@32 \\
\midrule
\emph{All prompts} & & & & & & \\
\quad $p_\theta$ & 0.59 & 0.83 & 0.91 & \textbf{0.82} & \textbf{0.91} & \textbf{0.94} \\
\quad $q_\phi$ & \textbf{0.71} & \textbf{0.96} & \textbf{0.99} & 0.78 & \textbf{0.91} & \textbf{0.94} \\
\midrule
\emph{Hard prompts ($p_\theta$ correct $\le 25\%$)} & & & & & & \\
\quad $p_\theta$ & 0.09 & 0.42 & 0.67 & 0.07 & 0.31 & 0.52 \\
\quad $q_\phi$ & \textbf{0.45} & \textbf{0.89} & \textbf{0.97} & \textbf{0.15} & \textbf{0.42} & \textbf{0.57} \\
\bottomrule
\end{tabular}
  \caption{Forward consistency rates of $q_\phi$ compared with the $p_\theta$ baseline (where FCR is just the pass rate) on all prompts and on the hard subset ($p_\theta$ correct at most 25\% of the time).}
  \label{table:fcr}
\end{table}

\subsection{Distillation from $q_\phi$ to $p_\theta$} \label{sec:distill-q-to-p}
Here we do two general experiments. In the first, to show the generalizability of $q_\phi$, we distill to $p_\theta$ using a dataset that $q_\phi$ was not trained on. This mimics the important real-world continual learning situation where new prompt-answer pairs arrive and we want to efficiently update $p_\theta$.

In the second experiment we distill on the same dataset that $q_\phi$ (and, via GRPO, $p_\theta$) was trained on. With the data held fixed, this experiment isolates how much additional learning $q_\phi$ can extract from a dataset on which GRPO performance has already saturated.

\subsubsection{Fresh data} \label{sec:fresh-data}
Because of the small dataset size of GSM8K and it only having a train/test split, for this experiment we only use the Countdown dataset. Let $\mathcal D_0$ be the dataset that $q_\phi$ was trained on and $\mathcal D_1$ be a fresh dataset (so $\mathcal D' = \mathcal D_1$ in the notation of Section \ref{sec:distill-q-to-p-method}). $\mathcal D_1$ consists of 40,000 Countdown problems generated with the same procedure and parameters as $\mathcal D_0$ but a different random seed, deduplicated against all splits of $\mathcal D_0$ so that no problem (as a multiset of numbers and target) appears in both. This dataset is used for training only and all evaluations are on $\mathcal D_0$'s test split. The up-front generation on $\mathcal D_1$ consists of 32 rollouts per prompt from $p_\theta$ (which supplies $\mathcal D'_{RS_{p_\theta}}$, the all-correct filter of Section \ref{sec:distill-q-to-p-method}, and the per-prompt difficulty estimates used in evaluation) and one answer-conditioned CoT per prompt from $q_\phi$. To study the effect of the mix ratio, we run distillation for each $\mu \in \{0, 0.25, 0.5, 0.75, 1\}$. Note that $\mu = 0$ is rejection sampling \citep{yuan2023scaling}, which we use as a baseline to compare against. For other baseline comparisons we use the performance of $p_\theta$ before distillation and of $p_\theta$ further finetuned, using GRPO, on $\mathcal D_1$.

We evaluate with pass@$k$, the probability that at least one of $k$ sampled completions answers correctly, and its restriction to the hard subset defined above, hard pass@$k$. For all runs in this experiment, we choose each run's checkpoint by its best hard pass@8 over $\mathcal D_0$'s val split, and then report all metrics from evaluating that checkpoint once on the held-out test split. We repeat this for each experiment with two seeds and report the mean and standard deviation in Table \ref{table:distill-exp1}.

Distillation with $\mu \ge 0.5$ is strongest on every pass@8 and pass@32 metric, while the baselines retain an edge at pass@1 (rejection sampling on all prompts, continued GRPO on the hard subset). This split is consistent with what each method optimizes: training only on the model's own reachable solutions (rejection sampling) or sharpening the current policy's modes (GRPO) improves single-sample precision, whereas distilling $q_\phi$'s traces adds newly solvable prompts whose solutions are initially sampled with low probability, a gain that surfaces at larger $k$.

\begin{table}[h]
  \centering
  \setlength{\tabcolsep}{3pt}
  \begin{tabular}{lllllll}
\toprule
& hard pass@1 & hard pass@8 & hard pass@32 & pass@1 & pass@8 & pass@32 \\
\midrule
mix ratio 1.0 & \pmval{0.19}{0.00} & \pmval{0.62}{0.00} & \pmval{0.85}{0.00} & \pmval{0.54}{0.00} & \pmval{0.87}{0.00} & $\mathbf{0.96}{\scriptstyle\,\pm0.00}$ \\
mix ratio 0.75 & \pmval{0.19}{0.00} & $\mathbf{0.64}{\scriptstyle\,\pm0.00}$ & $\mathbf{0.87}{\scriptstyle\,\pm0.00}$ & \pmval{0.57}{0.02} & $\mathbf{0.88}{\scriptstyle\,\pm0.00}$ & $\mathbf{0.96}{\scriptstyle\,\pm0.00}$ \\
mix ratio 0.5 & \pmval{0.18}{0.01} & \pmval{0.63}{0.02} & $\mathbf{0.87}{\scriptstyle\,\pm0.01}$ & \pmval{0.56}{0.02} & $\mathbf{0.88}{\scriptstyle\,\pm0.01}$ & $\mathbf{0.96}{\scriptstyle\,\pm0.00}$ \\
mix ratio 0.25 & \pmval{0.16}{0.01} & \pmval{0.59}{0.01} & \pmval{0.84}{0.00} & \pmval{0.60}{0.00} & \pmval{0.87}{0.00} & \pmval{0.95}{0.00} \\
mix ratio 0 (RS) & \pmval{0.13}{0.00} & \pmval{0.50}{0.02} & \pmval{0.75}{0.02} & $\mathbf{0.61}{\scriptstyle\,\pm0.01}$ & \pmval{0.85}{0.00} & \pmval{0.93}{0.01} \\
\midrule
GRPO full finetune & $\mathbf{0.23}{\scriptstyle\,\pm0.01}$ & \pmval{0.59}{0.02} & \pmval{0.79}{0.02} & \pmval{0.59}{0.00} & \pmval{0.85}{0.01} & \pmval{0.93}{0.01} \\\midrule
Base $p_\theta$ & 0.08 & 0.38 & 0.64 & 0.58 & 0.81 & 0.90 \\
\bottomrule
\end{tabular}
  \caption{Distillation on fresh Countdown data ($\mathcal D_1$). Rows with mix ratio $\mu > 0$ are our method; $\mu = 0$ is the rejection-sampling baseline, and we also compare against continued GRPO training on $\mathcal D_1$ and the starting model $p_\theta$. We report mean $\pm$ standard deviation over two seeds.}
  \label{table:distill-exp1}
\end{table}

\paragraph{Efficiency} Our method is clearly more data-efficient than traditional SFT techniques that require groundtruth CoT traces, and in this section we highlight the structural properties of it that make it more compute-efficient than policy gradient methods (such as GRPO) from a continual learning perspective, where the recurring cost is ingesting each new batch of prompt-answer data.

\emph{Generation cost scales with data, not with training.} Every completion GRPO learns from must be sampled from the current policy, inside the training loop: prompts are re-rolled-out at every visit, so generation cost grows with training length (and dominated it in our runs). Our pipeline computes rollouts for each batch exactly once, and the artifacts are reused by every subsequent run on that batch (e.g. all ten SFT runs behind Table \ref{table:distill-exp1} used the same rollout data). The inverse model itself is a one-time cost: $q_\phi$ was trained on $\mathcal D_0$ and applied to $\mathcal D_1$ with no retraining.

\emph{Generation decouples from training.} Because the SFT dataset is fixed by design, its rollouts never go stale: they can be produced on separate, inference-optimized hardware concurrently with training. Offline generation also batches continuously over the entire dataset at full GPU occupancy, whereas on-policy generation is capped at one group per prompt and drains as short completions finish, which leads to significantly less throughput.  

\emph{Training memory scales with one sequence, not a group.} A GRPO update computes gradients through the log-probabilities of every completion in a group, so group size is capped by training memory. SFT backpropagates through a single CoT per example, however many rollouts were sampled offline.

\emph{Compute converts into signal on the hard tail.} On the 17\% of fresh prompts that $p_\theta$ never solves, every GRPO rollout group is uniformly incorrect and the gradient is identically zero however long training runs, while $q_\phi$ produces forward-consistent CoT for 30\% of exactly those prompts. Consistently, continued GRPO plateaued below the $\mu \ge 0.5$ distillation arms on hard pass@8 and pass@32 (Table \ref{table:distill-exp1}) despite running for more than twice the RL that produced $p_\theta$.

\subsubsection{Original data} \label{sec:og-data}
\begin{figure}[t]
  \centering
  \begin{subfigure}{0.48\textwidth}
    \includegraphics[width=\linewidth]{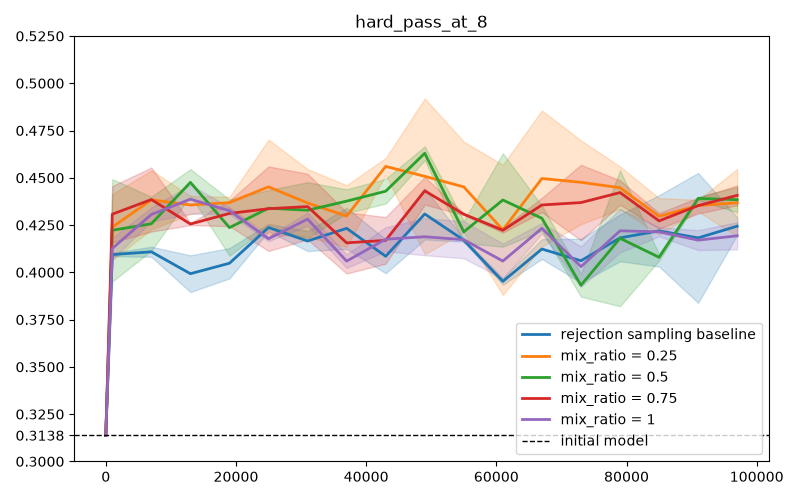}
    \caption{GSM8K, hard pass@8}
  \end{subfigure}
  \hfill
  \begin{subfigure}{0.48\textwidth}
    \includegraphics[width=\linewidth]{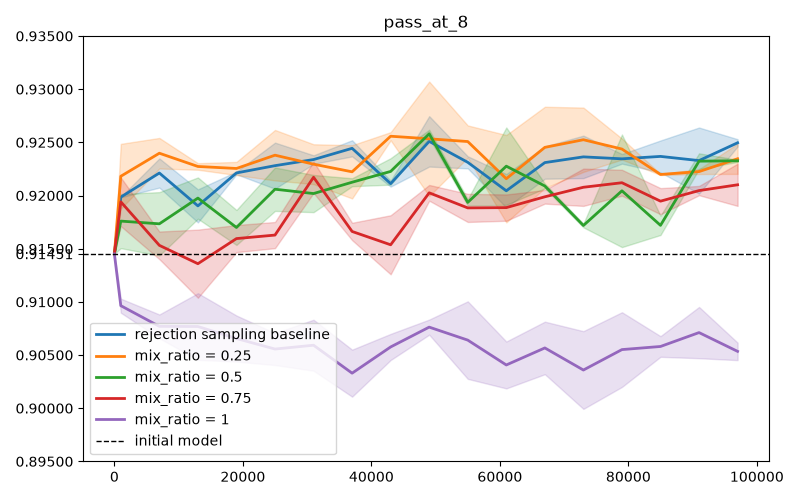}
    \caption{GSM8K, pass@8}
  \end{subfigure}

  \vspace{0.5em}

  \begin{subfigure}{0.48\textwidth}
    \includegraphics[width=\linewidth]{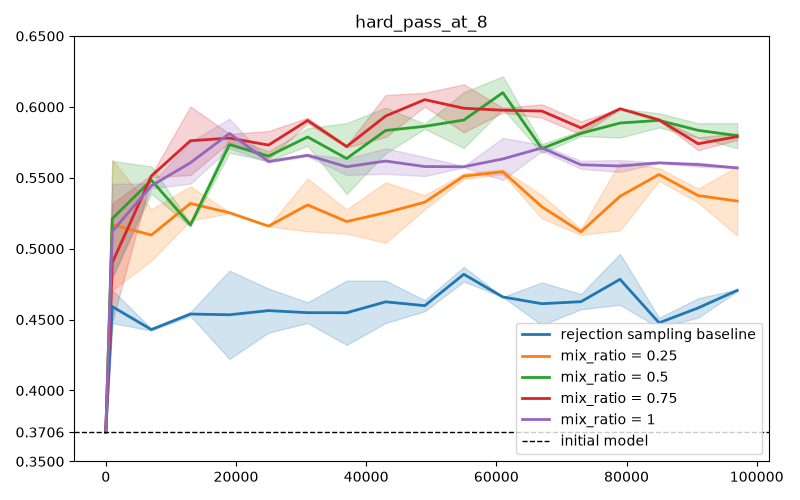}
    \caption{Countdown, hard pass@8}
  \end{subfigure}
  \hfill
  \begin{subfigure}{0.48\textwidth}
    \includegraphics[width=\linewidth]{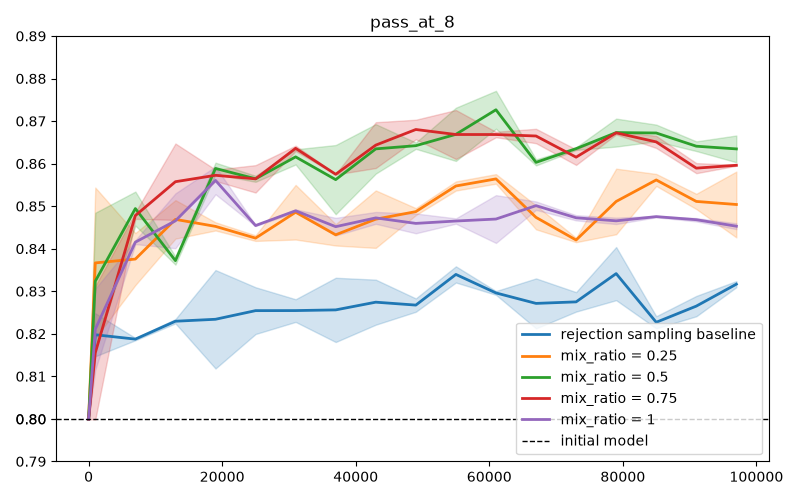}
    \caption{Countdown, pass@8}
  \end{subfigure}

  \caption{Held-out evaluation curves over training checkpoints on GSM8K (top)
  and Countdown (bottom), for each mix ratio $\mu$ and the rejection-sampling
  baseline ($\mu = 0$). Shaded bands show $\pm 1$ standard deviation across the
  two seeds; the dashed line is the initial model $p_\theta$. Curves for the
  remaining values of $k$ are in Appendix~\ref{appendix:additional-metrics}.}
  \label{fig:distill-exp2}
\end{figure}

These experiments are the same as above except we distill on $\mathcal D_0$, the dataset $q_\phi$ was trained on, and experiment on GSM8K as well as Countdown. For metrics we show held-out evaluation curves over training checkpoints (Figure \ref{fig:distill-exp2}) and report in Table \ref{table:distill-exp2} the mean over the last 30\% of checkpoints. On GSM8K the hard-tail gains over rejection sampling are smaller (at most $+0.03$ hard pass@32) and several cells tie, while pass@1 falls below both rejection sampling and the starting model at high mix ratios, the same precision-versus-coverage tradeoff seen on fresh data. Figure \ref{fig:qualitative-example-gsm8k} in Appendix \ref{appendix:qualitative} traces a held-out GSM8K prompt through this experiment where the starting $p_\theta$ fails on all 32 rollouts but the distilled model solves greedily, ruling out memorization.

\begin{table}[t]
  \centering
  \setlength{\tabcolsep}{3pt}
  \begin{tabular}{lllllll}
\toprule
& \multicolumn{6}{c}{GSM8K} \\
\cmidrule(lr){2-7}
& hard pass@1 & hard pass@8 & hard pass@32 & pass@1 & pass@8 & pass@32 \\
\midrule
mix ratio 1 & $\mathbf{0.15}{\scriptstyle\,\pm0.01}$ & \pmval{0.42}{0.00} & \pmval{0.62}{0.01} & \pmval{0.76}{0.00} & \pmval{0.91}{0.00} & \pmval{0.95}{0.00} \\
mix ratio 0.75 & \pmval{0.14}{0.00} & $\mathbf{0.44}{\scriptstyle\,\pm0.00}$ & $\mathbf{0.65}{\scriptstyle\,\pm0.00}$ & \pmval{0.79}{0.00} & $\mathbf{0.92}{\scriptstyle\,\pm0.00}$ & $\mathbf{0.96}{\scriptstyle\,\pm0.00}$ \\
mix ratio 0.5 & \pmval{0.13}{0.00} & \pmval{0.42}{0.01} & \pmval{0.63}{0.02} & \pmval{0.80}{0.00} & $\mathbf{0.92}{\scriptstyle\,\pm0.00}$ & \pmval{0.95}{0.00} \\
mix ratio 0.25 & \pmval{0.14}{0.00} & $\mathbf{0.44}{\scriptstyle\,\pm0.01}$ & \pmval{0.64}{0.00} & \pmval{0.80}{0.00} & $\mathbf{0.92}{\scriptstyle\,\pm0.00}$ & $\mathbf{0.96}{\scriptstyle\,\pm0.00}$ \\
mix ratio 0 (RS) & \pmval{0.13}{0.00} & \pmval{0.42}{0.00} & \pmval{0.62}{0.00} & \pmval{0.81}{0.00} & $\mathbf{0.92}{\scriptstyle\,\pm0.00}$ & \pmval{0.95}{0.00} \\
\midrule
Base $p_\theta$ & 0.07 & 0.31 & 0.53 & \textbf{0.82} & 0.91 & 0.94 \\
\bottomrule
\end{tabular}

  \vspace{1em}

  \begin{tabular}{lllllll}
\toprule
& \multicolumn{6}{c}{Countdown} \\
\cmidrule(lr){2-7}
& hard pass@1 & hard pass@8 & hard pass@32 & pass@1 & pass@8 & pass@32 \\
\midrule
mix ratio 1 & $\mathbf{0.18}{\scriptstyle\,\pm0.00}$ & \pmval{0.56}{0.00} & \pmval{0.79}{0.00} & \pmval{0.57}{0.00} & \pmval{0.85}{0.00} & \pmval{0.93}{0.00} \\
mix ratio 0.75 & \pmval{0.17}{0.00} & $\mathbf{0.59}{\scriptstyle\,\pm0.00}$ & $\mathbf{0.83}{\scriptstyle\,\pm0.00}$ & \pmval{0.58}{0.00} & \pmval{0.86}{0.00} & $\mathbf{0.95}{\scriptstyle\,\pm0.00}$ \\
mix ratio 0.5 & \pmval{0.16}{0.00} & \pmval{0.58}{0.00} & $\mathbf{0.83}{\scriptstyle\,\pm0.00}$ & \pmval{0.59}{0.00} & $\mathbf{0.87}{\scriptstyle\,\pm0.00}$ & $\mathbf{0.95}{\scriptstyle\,\pm0.00}$ \\
mix ratio 0.25 & \pmval{0.14}{0.00} & \pmval{0.53}{0.01} & \pmval{0.79}{0.01} & \pmval{0.58}{0.00} & \pmval{0.85}{0.00} & \pmval{0.94}{0.00} \\
mix ratio 0 (RS) & \pmval{0.12}{0.00} & \pmval{0.46}{0.00} & \pmval{0.72}{0.00} & $\mathbf{0.60}{\scriptstyle\,\pm0.00}$ & \pmval{0.83}{0.00} & \pmval{0.91}{0.00} \\
\midrule
Base $p_\theta$ & 0.08 & 0.37 & 0.62 & 0.57 & 0.80 & 0.88 \\
\bottomrule
\end{tabular}
  \caption{Distillation on the original data ($\mathcal D_0$) for GSM8K (top; evaluated on the standard held-out GSM8K test set) and Countdown (bottom; evaluated on the held-out test split). Values are the mean over the last 30\% of training checkpoints, reported as mean $\pm$ standard deviation over two seeds.}
  \label{table:distill-exp2}
\end{table}

\FloatBarrier

\subsection{$q_\phi$-verifier results} \label{sec:q-verifier-exp}
We now evaluate the second application of $q_\phi$, as a test-time verifier that selects among candidate completions of $p_\theta$. We do two experiments: a fixed-budget best-of-$n$ comparison and an adaptive variant \citep{aggarwal2023let} where sampling stops early once a confidence condition is met.

\subsubsection{Best-of-$n$}
For baselines we take the following:

\begin{itemize}
\item \textbf{Majority vote} \citep{wang2022self} From the $n$ samples return the answer that appeared most often.
\item \textbf{Shortest CoT} Many incorrect responses, especially for Countdown, have long CoT, as the model has to try a bunch of different ideas from scratch trying to get the correct answer. As such, we consider the selector that from the $n$ samples returns the one with shortest CoT sequence. 
\item \textbf{$p_\theta$'s self-confidence} The log probs of $p_\theta$'s completion of the prompt.
\end{itemize}

Our results are in Table \ref{table:best-of-n}. We restrict to \emph{mixed} prompts, i.e. those whose $n$ samples contain both a correct and an incorrect answer, since on the remaining prompts every selector scores identically and including them would only dilute the comparison. On Countdown, selecting with $q_\phi$ cuts the selection error of majority vote from 24.7\% to 3.1\%. On GSM8K, the aggregate difference between $q_\phi$ and majority vote (0.765 vs.\ 0.754 on 472 prompts) is within noise at this sample size; Table \ref{table:difficulty-strata} shows where they differ. Stratifying by whether the correct completions form a minority of the samples (i.e. whether $p_\theta$'s consensus is wrong), we see that $q_\phi$'s advantage concentrates precisely on the correct-minority prompts, where majority vote is by construction at or below random. This is the same hard-tail regime targeted by our distillation experiments, and exactly where a verifier is most valuable. On correct-majority prompts voting is near-perfect and $q_\phi$ adds nothing (slightly losing to it on GSM8K).

\begin{table}[t]
  \begin{minipage}[t]{0.47\textwidth}
    \centering
    \small
    \setlength{\tabcolsep}{3.5pt}
    \begin{tabular}{lcc}
\toprule
Selector & Countdown & GSM8K \\
\midrule
$q_\phi$ & \textbf{0.969} & \textbf{0.765} \\
majority vote & 0.753 & 0.754 \\
shortest CoT & 0.869 & 0.699 \\
$p_\theta$'s self-confidence & 0.757 & 0.631 \\
random & 0.561 & 0.638 \\
\bottomrule
\end{tabular}

    \caption{Selection accuracy of best-of-$n$ techniques for $n=32$, on mixed prompts (those with at least one correct and one incorrect sample; 3,483 for Countdown, 472 for GSM8K).}
    \label{table:best-of-n}
  \end{minipage}
  \hfill
  \begin{minipage}[t]{0.49\textwidth}
    \centering
    \small
    \setlength{\tabcolsep}{3.5pt}
    \begin{tabular}{lcc}
\toprule
Selector & Countdown & GSM8K \\
\midrule
\emph{Correct-minority} & \emph{($n$ = 1,387)} & \emph{($n$ = 141)} \\
\quad $q_\phi$ & \textbf{0.929} & \textbf{0.383} \\
\quad majority vote & 0.388 & 0.193 \\
\quad random & 0.196 & 0.217 \\
\midrule
\emph{Correct-majority} & \emph{($n$ = 2,096)} & \emph{($n$ = 331)} \\
\quad $q_\phi$ & \textbf{0.995} & 0.927 \\
\quad majority vote & 0.991 & \textbf{0.997} \\
\quad random & 0.816 & 0.848 \\
\bottomrule
\end{tabular}

    \caption{Best-of-$n$ selection accuracy stratified by whether correct completions are a minority among the prompt's samples, i.e.\ whether $p_\theta$'s consensus is wrong.}
    \label{table:difficulty-strata}
  \end{minipage}
\end{table}


\subsubsection{Adaptive best-of-$n$}
Adaptive best-of-$n$ techniques \citep{aggarwal2023let} involve the specification of a stopping criterion so that an answer may be returned before generating all $n$ completions. A threshold $\tau$ on the verifier determines such a condition by returning the first generated answer with $q_\phi(C | P, A) > \tau$. For baselines, we take adaptive analogs of majority vote and of the shortest-CoT selector: for a positive integer $m$, we stop as soon as the most common answer appears $m$ more times than any other; and for a length $L$, we return the first completion whose CoT has at most $L$ tokens (falling back to the shortest seen). To make a fair comparison, we sweep $\tau$, $m$, and $L$ respectively and, for each setting, measure both the selection accuracy and the expected number of completions drawn per prompt, using the same random completion orderings for all three. Table \ref{table:adaptive-budget} compares the methods at matched expected budgets.

On Countdown, $q_\phi$ at ${\sim}2$ expected draws already exceeds adaptive majority vote at \emph{any} budget. The length rule is the stronger baseline on Countdown, beating the vote margin at every budget, yet $q_\phi$ leads it by 6--13 points throughout, so the advantage is not a length artifact; on GSM8K the length rule is weak, consistent with Table \ref{table:best-of-n}. Against majority vote on GSM8K a significant advantage is confined to small budgets. The structural reason is that a vote margin is a population signal, in particular it cannot trigger before $m$ draws and may never build up when answers scatter. However a verifier score is a per-sample signal that can trigger after a single draw.

\begin{table}[t]
  \centering
  \begin{tabular}{lcccccc}
\toprule
 & \multicolumn{3}{c}{Countdown} & \multicolumn{3}{c}{GSM8K} \\
\cmidrule(lr){2-4} \cmidrule(lr){5-7}
Expected draws & $q_\phi$ & margin & length & $q_\phi$ & margin & length \\
\midrule
$\sim2$ & \textbf{0.767} & 0.606 & 0.705 & \textbf{0.747} & 0.694 & 0.688 \\
$\sim4$ & \textbf{0.962} & 0.667 & 0.835 & \textbf{0.752} & 0.746 & 0.698 \\
$\sim8$ & \textbf{0.968} & 0.740 & 0.888 & \textbf{0.762} & 0.756 & 0.700 \\
full $N$ ($\sim$32) & \textbf{0.969} & 0.753 & 0.869 & \textbf{0.765} & 0.754 & 0.699 \\
\bottomrule
\end{tabular}

  \caption{Adaptive best-of-$n$: selection accuracy at matched expected numbers of completions drawn per prompt, for the $q_\phi$-threshold stopping rule versus the adaptive majority-vote (margin) and length-threshold (length) baselines, on mixed prompts.}
  \label{table:adaptive-budget}
\end{table}

\FloatBarrier

\section{Conclusion and future work}

In this work we demonstrated the utility of treating CoT tokens as latent variables and training an inverse model $q_\phi$ that generates reasoning conditioned on both the prompt and groundtruth answer. Our experiments show that after RL training has saturated, cheap SFT training using $q_\phi$-generated traces recovers substantial performance on the hard tail of prompts. These are problems that rejection sampling discards entirely because the forward model never solves them. Further, the effectiveness of this SFT technique extends to fresh answer-only data, making the approach suitable as an offline, continual-learning update that avoids the per-step rollout cost of policy-gradient methods such as GRPO. We have also shown that $q_\phi$ is a suitable verifier for correctness, enabling efficient best-of-$n$ and adaptive test-time selection precisely in the regime where majority voting fails.

In terms of sample efficiency, our method occupies a middle ground between RL and SFT: the mixed correct/incorrect rollouts that GRPO needs at every training step are needed only once, up front, to train $q_\phi$ via the contrastive objective. All subsequent learning is supervised. We also note that little in the method is specific to LLMs or to CoT: the same recipe applies to any latent-variable sequence model where an observed outcome can be conditioned on to propose the latent.

For future work, a natural extension is to iterate: refresh $q_\phi$ against the updated $p_\theta$ and alternate the two updates, which places the method in the expectation-maximization/wake-sleep family, with the single-step version presented here as one outer step. Relatedly, generating multiple CoT candidates per prompt would permit self-normalized importance weighting in place of our hard forward-consistency filter. Finally, scaling the study to larger models and harder environments, and using $q_\phi$-synthesized traces to re-seed RL on the hard tail where GRPO currently receives no gradient signal, are both promising directions.

\bibliography{refs}
\bibliographystyle{colm2026_conference}

\appendix
\raggedbottom 
\section{Training details} \label{appendix:training-details}
\subsection{GRPO training of $p_\theta$} \label{sec:grpo-training-details}
Table \ref{table:grpo-hyperparams} summarizes the hyperparameters used for GRPO. They are shared by the initial training of $p_\theta$ on each environment and by the continued-GRPO baseline of the fresh-data experiment; the latter is an exact continuation of the original run---same hyperparameters, with optimizer state restored---on the fresh dataset $\mathcal D_1$. Unlike $q_\phi$ and the distillation SFT below, GRPO updates all model weights: in preliminary experiments we found GRPO training with LoRA to be unstable.

\begin{table}[H]
  \centering
  \begin{tabular}{lc}
\toprule
Hyperparameter & Value \\
\midrule
\emph{GRPO} & \\
\quad Group size $g$ & 8 \\
\quad KL weight $\beta$ & 0 \\
\quad Advantage normalization & Yes \\
\quad Length normalization & Yes \\
\quad Reward & correctness $+\ 0.1 \times$ answer-format \\
\midrule
\emph{Optimization} & \\
\quad Finetuning & full (no LoRA) \\
\quad Optimizer & Adam \\
\quad Learning rate & $1\times10^{-5}$ \\
\quad Prompts per step & 32 \\
\quad Gradient clipping & 1 \\
\quad Precision & bf16 \\
\midrule
\emph{Sampling} & \\
\quad Temperature & 1 \\
\quad Max CoT tokens & 600 \\
\bottomrule
\end{tabular}

  \caption{Hyperparameters used for GRPO training of $p_\theta$ (both the initial training and the continued-GRPO baseline).}
  \label{table:grpo-hyperparams}
\end{table}

\subsection{Training $q_\phi$} \label{sec:q-training-details}
The hyperparameters used to train $q_\phi$ are summarized in Table \ref{table:q-hyperparams}. Here $m$ is the margin and $\lambda$ the weight of the contrastive loss term (see eq. \ref{eq:contrastive}).

\begin{table}[H]
  \centering
  \begin{tabular}{lc}
\toprule
Hyperparameter & Value \\
\midrule
\emph{Contrastive loss} & \\
\quad Margin $m$ & 1 \\
\quad Loss weight $\lambda$ & 1 \\
\quad Normalize by sequence length & Yes \\
\midrule
\emph{LoRA} & \\
\quad Rank $r$ & 32 \\
\quad $\alpha$ & 64 \\
\midrule
\emph{Optimization} & \\
\quad Optimizer & Adam \\
\quad Learning rate & $2\times10^{-4}$ \\
\quad Batch size & 32 \\
\quad Weight decay & 0.01 \\
\quad Gradient clipping & 10 \\
\quad Precision & bf16 \\
\midrule
\emph{Sampling} & \\
\quad Temperature & 1 \\
\quad Max CoT tokens & 600 \\
\quad Group size $g$ & 32 \\
\bottomrule
\end{tabular}

  \caption{Hyperparameters used for training $q_\phi$.}
  \label{table:q-hyperparams}
\end{table}

We first train $q_\phi$ for Countdown and saw FCR scores on the val split plateau around optimization step 25,000. When training the smaller dataset GSM8K, which only has a train/test split, we therefore did not do any train-time validation and simply took the checkpoint at step 25,000 for $q_\phi$.

\subsection{Distilling $q_\phi$ into $p_\theta$} \label{sec:sft-training-details}
The hyperparameters used for the distillation SFT are summarized in Table \ref{table:sft-hyperparams}, and are shared across both the Countdown and GSM8K experiments. The only hyperparameter we vary is the mix ratio $\mu$, which controls the fraction of CoTs sampled from $q_\phi$ versus $p_\theta$.

\begin{table}[H]
  \centering
  \begin{tabular}{lc}
\toprule
Hyperparameter & Value \\
\midrule
\emph{LoRA} & \\
\quad Rank $r$ & 16 \\
\quad $\alpha$ & 32 \\
\midrule
\emph{Optimization} & \\
\quad Optimizer & Adam \\
\quad Learning rate & $5\times10^{-4}$ \\
\quad Batch size & 16 \\
\quad Gradient clipping & 1 \\
\quad Precision & bf16 \\
\midrule
\emph{Distillation} & \\
\quad Mix ratio $\mu$ & varied \\
\quad Max CoT tokens & 600 \\
\quad Temperature & 1 \\
\bottomrule
\end{tabular}

  \caption{Hyperparameters used for distilling $q_\phi$ into $p_\theta$.}
  \label{table:sft-hyperparams}
\end{table}

\FloatBarrier

\section{Additional metrics} \label{appendix:additional-metrics}
\begin{figure}[H]
  \centering
  \begin{subfigure}{0.48\textwidth}
    \includegraphics[width=\linewidth]{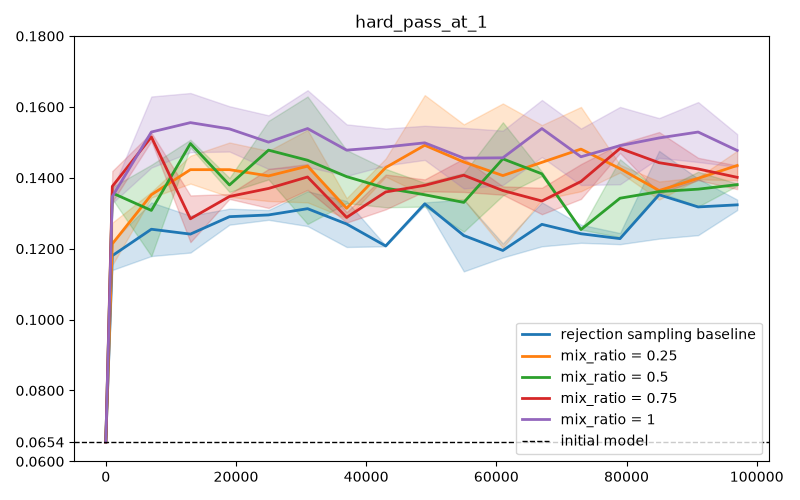}
    \caption{hard pass@1}
  \end{subfigure}
  \hfill
  \begin{subfigure}{0.48\textwidth}
    \includegraphics[width=\linewidth]{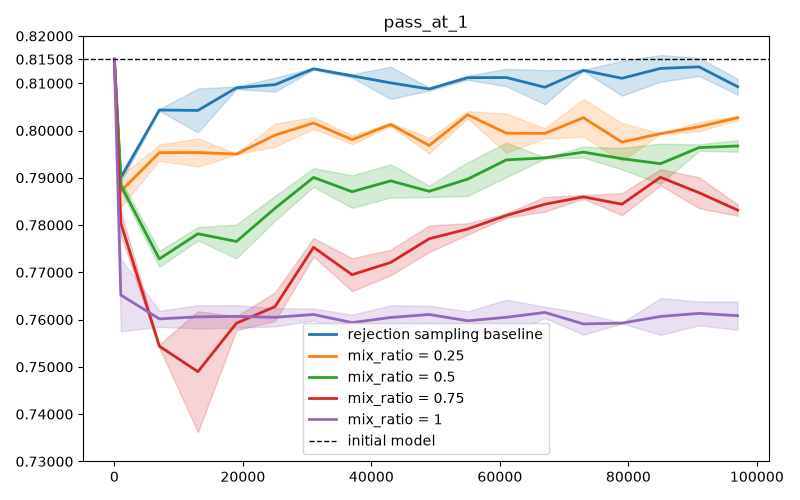}
    \caption{pass@1}
  \end{subfigure}

  \vspace{0.5em}

  \begin{subfigure}{0.48\textwidth}
    \includegraphics[width=\linewidth]{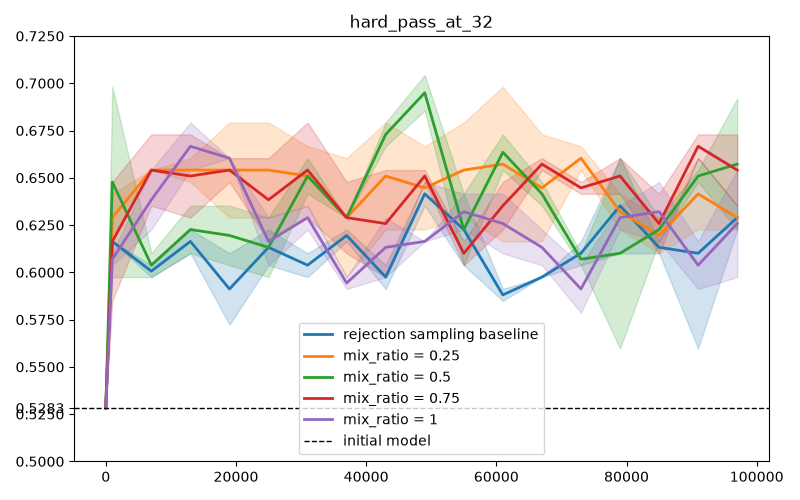}
    \caption{hard pass@32}
  \end{subfigure}
  \hfill
  \begin{subfigure}{0.48\textwidth}
    \includegraphics[width=\linewidth]{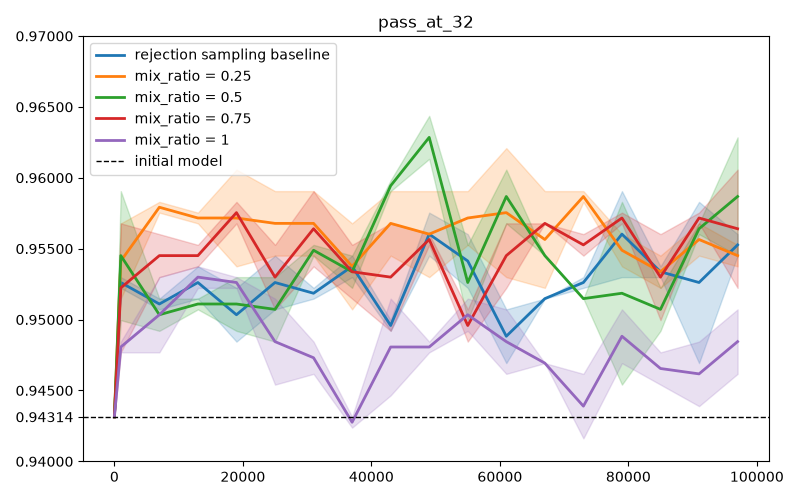}
    \caption{pass@32}
  \end{subfigure}

  \caption{Held-out evaluation curves over training checkpoints on GSM8K at
  $k \in \{1, 32\}$, complementing Figure~\ref{fig:distill-exp2}. Shaded
  bands show $\pm 1$ standard deviation across the two seeds; the dashed line is
  the initial model $p_\theta$.}
  \label{fig:distill-exp2-gsm8k-appendix}
\end{figure}

\begin{figure}[H]
  \centering
  \begin{subfigure}{0.48\textwidth}
    \includegraphics[width=\linewidth]{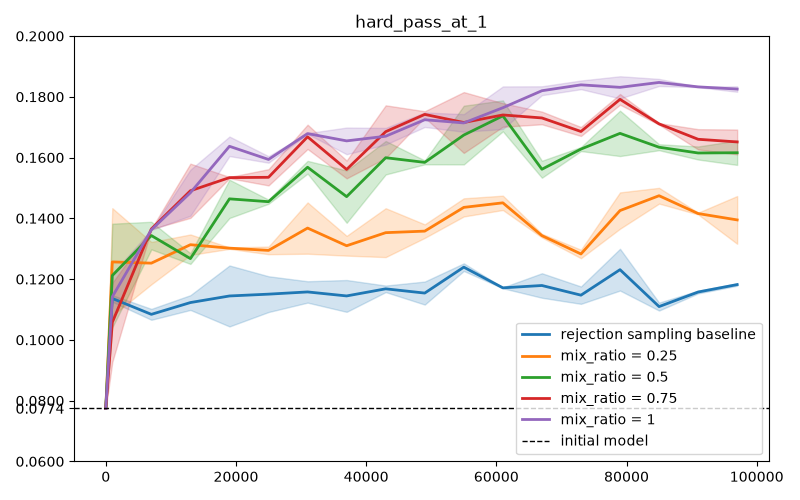}
    \caption{hard pass@1}
  \end{subfigure}
  \hfill
  \begin{subfigure}{0.48\textwidth}
    \includegraphics[width=\linewidth]{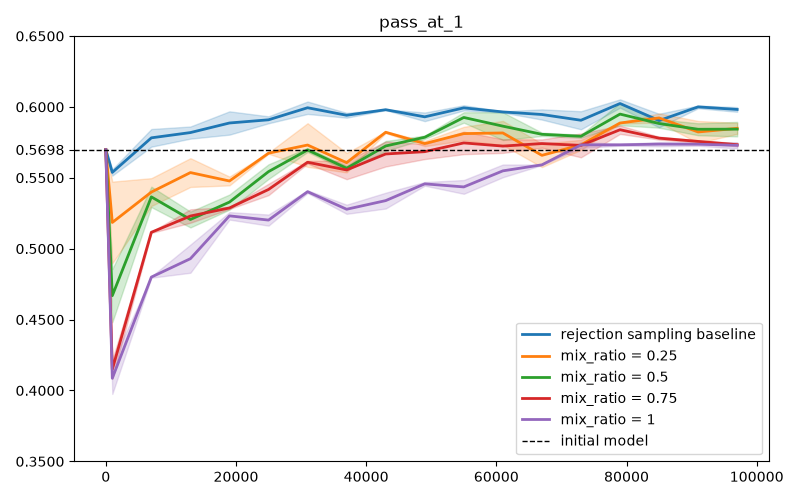}
    \caption{pass@1}
  \end{subfigure}

  \vspace{0.5em}

  \begin{subfigure}{0.48\textwidth}
    \includegraphics[width=\linewidth]{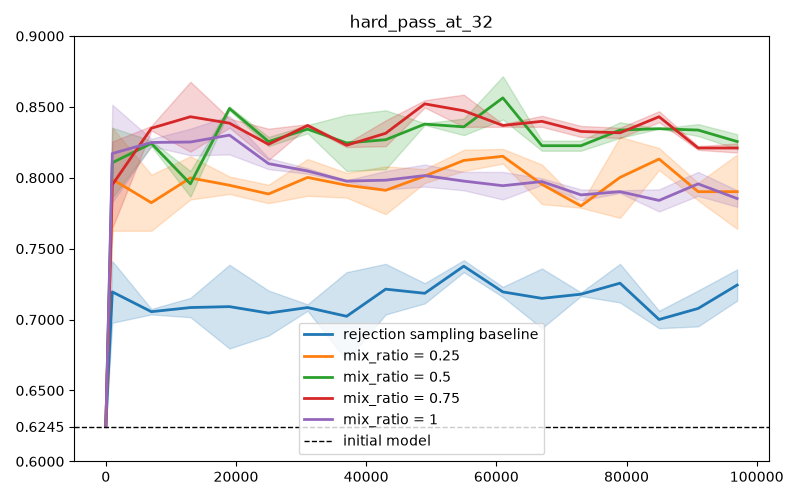}
    \caption{hard pass@32}
  \end{subfigure}
  \hfill
  \begin{subfigure}{0.48\textwidth}
    \includegraphics[width=\linewidth]{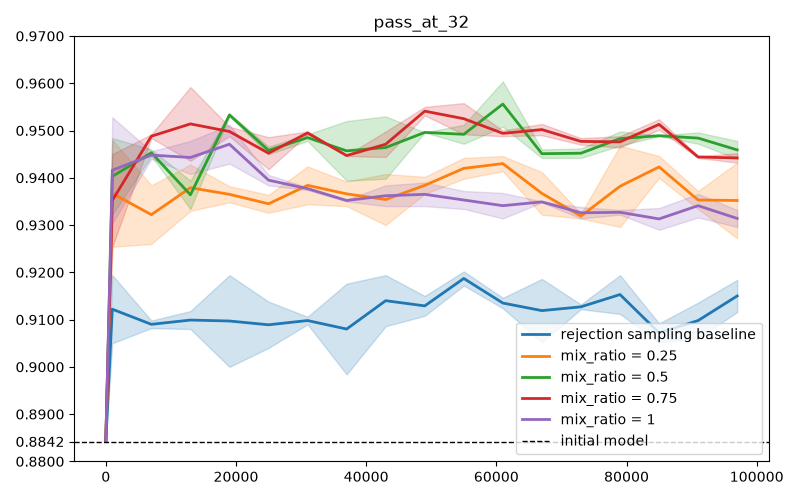}
    \caption{pass@32}
  \end{subfigure}

  \caption{Held-out evaluation curves over training checkpoints on Countdown at
  $k \in \{1, 32\}$, complementing Figure~\ref{fig:distill-exp2}. Shaded
  bands show $\pm 1$ standard deviation across the two seeds; the dashed line is
  the initial model $p_\theta$.}
  \label{fig:distill-exp2-countdown-appendix}
\end{figure}

Figures \ref{fig:distill-exp2-gsm8k-appendix} and \ref{fig:distill-exp2-countdown-appendix} show the held-out evaluation curves of the original-data distillation experiment for the values of $k$ not shown in the main text.

\FloatBarrier

\section{Qualitative examples} \label{appendix:qualitative}
Figures \ref{fig:qualitative-example} and \ref{fig:qualitative-example-gsm8k} each trace a single hard prompt through our pipeline: a prompt the starting $p_\theta$ never solves (0 of 32 rollouts), on which $q_\phi$ nevertheless produces a forward-consistent CoT, and which the distilled $p_\theta$ then solves with greedy decoding. The Countdown example (Figure \ref{fig:qualitative-example}) is a training prompt from the fresh-data experiment (Section \ref{sec:fresh-data}) and illustrates the transfer mechanism; the GSM8K example (Figure \ref{fig:qualitative-example-gsm8k}) is a \emph{held-out} prompt from the original-data experiment (Section \ref{sec:og-data}), so the distilled model's success cannot be memorization.

\begin{figure}[H]
\small
\begin{minipage}{\linewidth}
\textbf{Prompt (train):} Using the numbers [75, 7, 7, 4], create an equation that equals 244. You can use basic arithmetic operations (+, -, *, /) and each number exactly once. \emph{(gold: \texttt{4 * ((75 - 7) - 7)})}
\end{minipage}
\begin{itemize}
  \item[(a)] \textbf{$p_\theta$ before distillation} (all 32 rollouts incorrect; the shortest is shown). The arithmetic is correct, but the expression uses three 7s---a constraint violation that looks right:
  \begin{quote}\ttfamily\frenchspacing
  First, 244 divided by 7 is about 34.57, not helpful. How about (75 * 4) = 300, then minus 56. How get 56 from 7 and 7: 7*7=49, plus 7=56. So (75 * 4) - (7*7 + 7) = 300 - 56 = 244.
  \end{quote}

  \item[(b)] \textbf{$q_\phi$'s CoT}, conditioned on the groundtruth answer \mbox{\texttt{4 * ((75 - 7) - 7)}} (forward-consistent, and the actual SFT training datum). It decomposes backward from the target:
  \begin{quote}\ttfamily\frenchspacing
  First, 244 divided by 4 is 61. How do I get 61 from the other numbers? 75 - 14 = 61, but where do I get 14 from? 7 + 7 = 14. So (75 - (7 + 7)) = 75 - 14 = 61, then multiply by 4: 61*4=244.
  \end{quote}

  \item[(c)] \textbf{$p_\theta$ after distillation}, greedy decoding (deterministic, and with no answer conditioning), correct:
  \begin{quote}\ttfamily\frenchspacing
  <think> Okay, let's see. I need to get 244 using the numbers 75, 7, 7, and 4. Hmm. Let me think. First, 244 divided by 4 is 61. How can I get 61 from the other numbers: 75, 7, 7. Let's see. 75 - 7 = 68, minus 7 is 61. So (75 - (7 + 7)) - 4 = 75 - 14 = 61, then 61 * 4 = 244. </think> <answer> (75 - (7 + 7)) * 4 </answer>
  \end{quote}
\end{itemize}
\caption{A qualitative example from the fresh-data Countdown experiment. The starting model $p_\theta$ fails on all 32 rollouts (a); given the answer, $q_\phi$ derives a plan backward from the target---divide by 4, then decompose 61 (b); after distillation, $p_\theta$ executes the same divide-the-target strategy unaided (c). This is a training prompt, so (b) is the \emph{actual SFT training datum}, letting us see the supervision and what was learned from it side by side. Notably, even though the model trained on this very prompt, (c) is not a replay of (b): the student reaches 61 by a different route ($75 - 7 - 7$ rather than $75 - (7+7)$), slips and self-corrects (``- 4'' before ``* 4''), and brackets the final expression differently---what transfers is the strategy, not the memorized sequence. Held-out generalization is established quantitatively in Table~\ref{table:distill-exp1} and qualitatively in Figure~\ref{fig:qualitative-example-gsm8k}. All model outputs are kept verbatim.}
\label{fig:qualitative-example}
\end{figure}

\begin{figure}[H]
  \input{tables/distill/trio-example-gsm8k-heldout}
  \caption{A held-out GSM8K example from the original-data experiment. The starting $p_\theta$ fails all 32 rollouts, unanimously answering 6---correct arithmetic for the number of \emph{additional} hours, but not the total the question asks for (a). Conditioned on the gold answer 9, $q_\phi$'s derivation is forced through the step $p_\theta$ omits, with visible self-correction (b); its forward-consistency rate on this prompt (1 of 8) is in line with the hard-prompt FCR@1 of 0.15 in Table \ref{table:fcr}. The distilled $p_\theta$ solves the prompt greedily (c). Since the prompt is held out it contributed no training data---(b) is illustrative of the training signal on similar prompts---so memorization is impossible by construction. Outputs abridged where marked ([\ldots]).}
  \label{fig:qualitative-example-gsm8k}
\end{figure}

\end{document}